\documentclass{article}

\makeatletter
\def\input@path{{styles/}}
\makeatother
\usepackage[preprint]{colm2026_conference}
\usepackage{fontspec}
\usepackage{fontawesome5}

\renewcommand{\encodingdefault}{T1}
\normalfont
\usepackage{microtype}
\usepackage{graphicx}
\usepackage{trimclip}
\usepackage{float}
\usepackage{wrapfig}
\usepackage{amsmath}
\usepackage{amssymb}
\usepackage{booktabs}
\usepackage{multirow}
\usepackage{colortbl}

\usepackage{algorithm}
\usepackage{algpseudocode}
\usepackage{tikz}
\usepackage{tcolorbox}
\usepackage{hyperref}
\usepackage{url}
\usepackage{arydshln}
\usepackage{arydshln}
\definecolor{abyss}{HTML}{121D36}
\definecolor{electricblue}{HTML}{3866FF}
\definecolor{covercream}{HTML}{EEF3FA}
\definecolor{coveraccent}{HTML}{3866FF}

\newfontfamily\outfit[
  Path=assets/fonts/,
  UprightFont=Outfit-Regular.ttf,
  BoldFont=Outfit-SemiBold.ttf
]{Outfit}

\hypersetup{
  colorlinks=true,
  linkcolor=electricblue,
  citecolor=electricblue,
  urlcolor=coveraccent,
  filecolor=electricblue
}

\setcitestyle{numbers,square,comma,sort&compress}

\usepackage{amsmath,amsfonts,bm}

\def\eqref#1{equation~\ref{#1}}

\def\1{\bm{1}}

\DeclareMathAlphabet{\mathsfit}{\encodingdefault}{\sfdefault}{m}{sl}
\SetMathAlphabet{\mathsfit}{bold}{\encodingdefault}{\sfdefault}{bx}{n}

\newcommand{\reporttitle}{Verify Before You Distill:
Prompt-Level Teacher Gating for On-Policy Distillation}
\title{\reporttitle}

\author{AllSpark Team}

\begin{document}

\fancyhead{}
\renewcommand{\headrulewidth}{0pt}
\color{abyss}
\thispagestyle{empty}

\vspace*{-0.44in}
\begin{tcolorbox}[
  width=\linewidth,
  colback=covercream,
  colframe=covercream,
  boxrule=0pt,
  arc=14pt,
  outer arc=14pt,
  boxsep=0pt,
  left=20pt,
  right=20pt,
  top=13pt,
  bottom=11pt
]
  {\outfit\fontsize{21.5}{25.5}\selectfont\bfseries\centering
    \reporttitle\par}
  \vspace{1.25em}
  {\bfseries\centering AllSpark Team\par}

  \vspace{0.7em}
  \begingroup
  \normalfont
  \setlength{\parindent}{0pt}
  \setlength{\parskip}{0pt}
  On-policy distillation (OPD) accelerates post-training by providing dense token-level
supervision from a frozen teacher on the student's own rollouts. Vanilla OPD applies this
supervision uniformly across prompts, without checking whether the teacher is reliable for
each prompt. Because reverse KL is mode-seeking, a confidently wrong teacher can induce a
strong yet misleading update. Distributional proxies, such as entropy or teacher--student
likelihood agreement, measure uncertainty or agreement but do not directly verify outcome
correctness. We introduce Teacher-Gated On-Policy Distillation (TGOPD), built on the
principle that teacher reliability should be verified at the prompt level before dense
supervision is admitted. TGOPD estimates reliability from a small set of verifier-scored
teacher probes and routes each prompt exclusively to dense OPD when the reliability check
passes or to verifier-grounded GRPO otherwise. Across 4B and 35B students in mathematics,
code, and instruction following, TGOPD outperforms Vanilla OPD in all six single-domain
settings and achieves higher seven-benchmark averages at both scales under multi-domain
training. By using otherwise-idle teacher capacity for reliability estimation, TGOPD also
reduces teacher-side compute waste in asynchronous OPD, increasing teacher-node GPU
utilization from $9.8\%$ to $78.9\%$ in the measured 4B single-domain run.
  \par
  \endgroup

  \vspace{0.6em}
  \noindent
  \begin{minipage}[b]{0.63\linewidth}
    \outfit\fontsize{8.4}{10.2}\selectfont
    \textbf{Date:} August 28, 2026\\[-0.1em]
    \textbf{Status:} Public Technical Report
  \end{minipage}%
  \hfill
  \begin{minipage}[b]{0.33\linewidth}
    \raggedleft
    \raisebox{-0.30em}{\includegraphics[height=16pt]{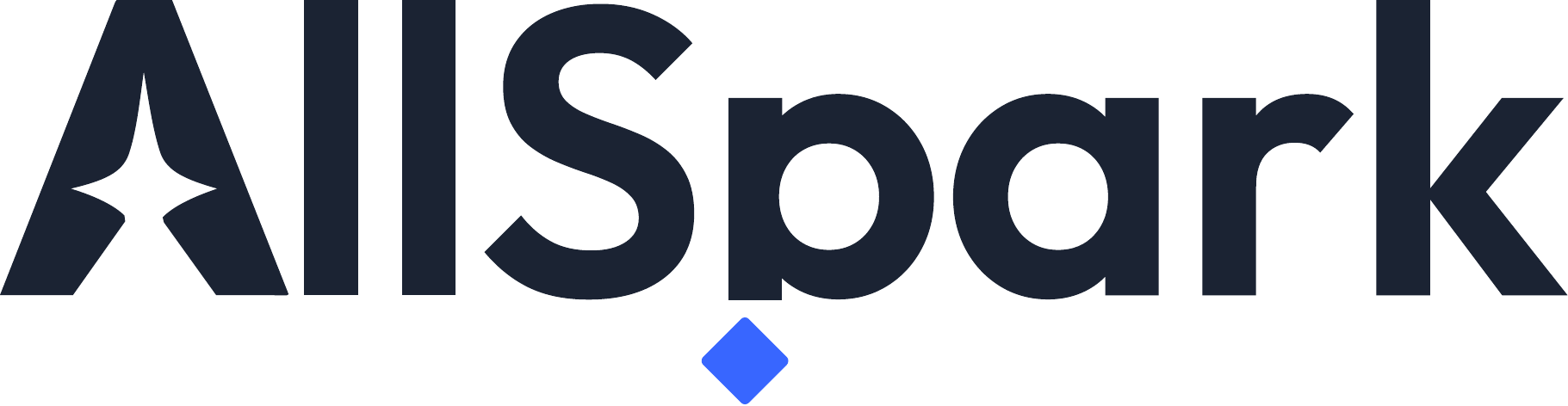}}%
  \end{minipage}
\end{tcolorbox}

\vspace{0.35em}

\section{Introduction}
\label{sec:intro}

\begin{wrapfigure}{r}{0.4\linewidth}
  \centering
  \vspace{-7pt}
  \includegraphics[width=0.97\linewidth]{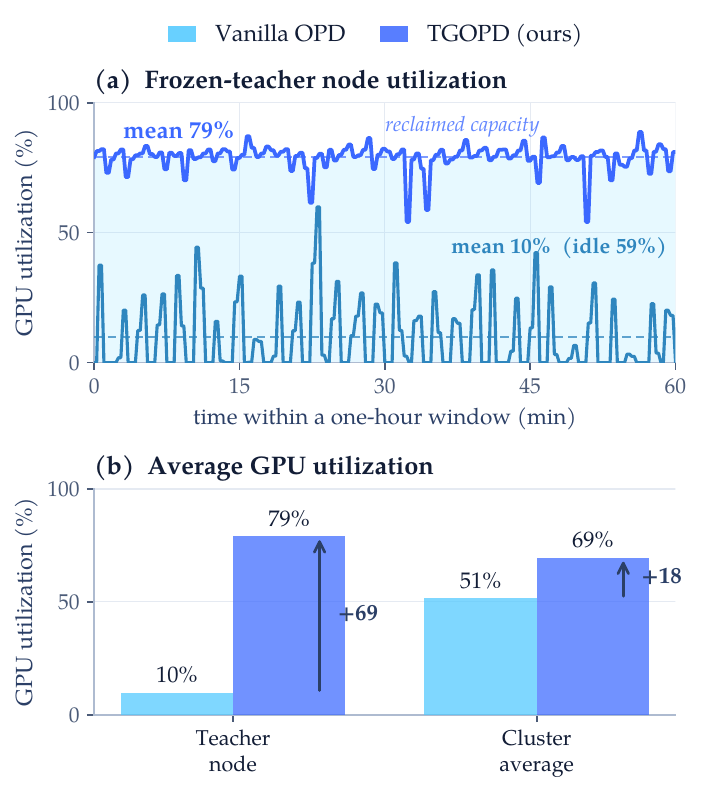}
  \vspace{-3pt}
  \caption{\textbf{Teacher-side GPU underutilization in asynchronous OPD.}
  \textbf{(a)} Under Vanilla OPD, teacher-node utilization remains below $5\%$
  for $59\%$ of the measured hour; TGOPD uses this otherwise-idle capacity for
  reliability probes.
  \textbf{(b)} Mean teacher-node utilization increases from $9.8\%$ to $78.9\%$.}
  \label{fig:idle}
  \vspace{-5pt}
\end{wrapfigure}

On-policy distillation (OPD) is a compute-efficient approach to post-training language
models \citep{agarwal2024gkd,gu2024minillm,thinkingmachines2025opd}.
Reinforcement learning with verifiable rewards (RLVR) assigns a single scalar reward to an
entire trajectory. In contrast, OPD samples rollouts from the student and uses a stronger
teacher to provide a reverse-KL learning signal at every token, transforming sparse outcome
supervision into dense per-token guidance. This dense feedback can enable the student to
reach teacher-level accuracy roughly an order of magnitude faster than RLVR
\citep{thinkingmachines2025opd}.

Despite this training efficiency, asynchronous OPD leaves much of the teacher's compute
idle. In a typical asynchronous deployment, the student generates rollouts
on dedicated inference nodes, while a stronger, domain-specialized frozen teacher scores
completed rollouts on a separate node. A scoring pass requires only a forward evaluation
over tokens that have already been generated. It is therefore much cheaper than
autoregressive decoding and cannot begin until a batch of student rollouts is ready,
leaving the teacher's accelerators idle between scoring batches.
Figure~\ref{fig:idle} quantifies this inefficiency in a 4B run: the teacher node averages
$9.8\%$ GPU utilization and remains below $5\%$ utilization for $59\%$ of the measured
hour, while the rollout and training nodes remain busy. This observation motivates using
the otherwise-idle teacher capacity to improve the training signal itself.

\begin{figure}[t]
  \centering
  \includegraphics[width=\linewidth]{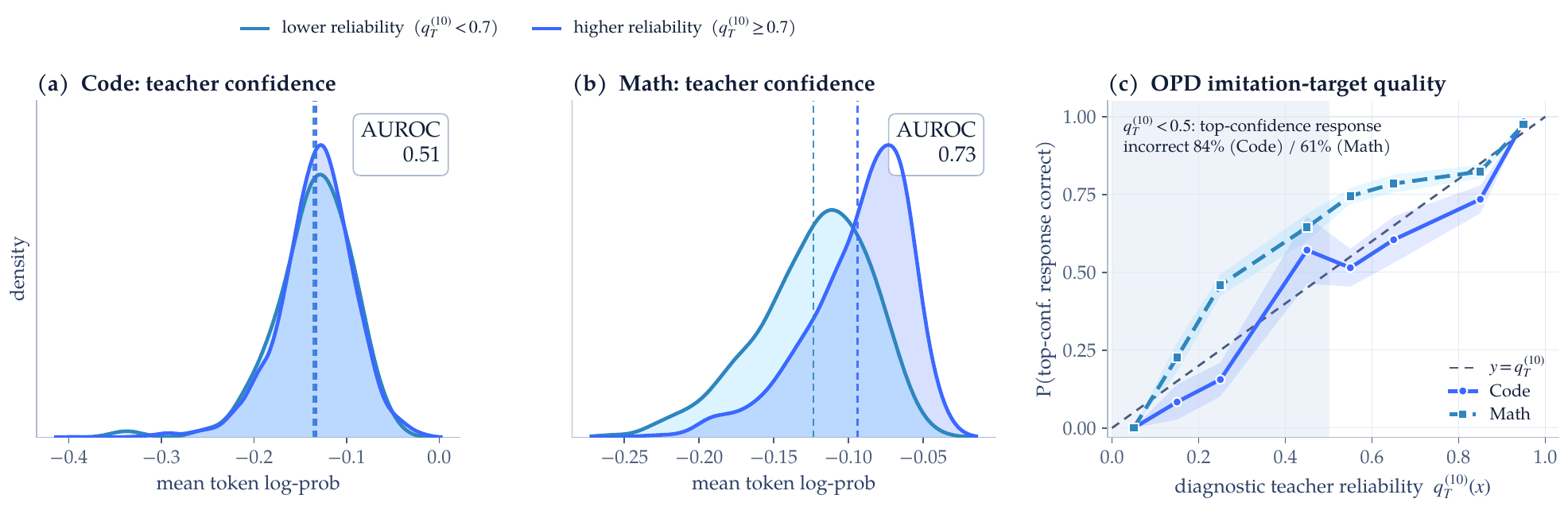}
  \caption{\textbf{Teacher confidence does not consistently indicate prompt-level
  reliability.}
  This diagnostic uses ten offline teacher responses for each of $2{,}400$ prompts per
  domain; $q_T^{(10)}$ is their verifier pass rate and is distinct from the $K_T=3$ online
  estimator used for training. \textbf{(a,b)} Teacher self-confidence under higher
  ($q_T^{(10)}\!\ge\!0.7$) and lower ($q_T^{(10)}\!<\!0.7$) reliability. Confidence separates
  the groups at AUROC~$0.51$ on code and $0.73$ on math. \textbf{(c)} In the low-reliability
  regime ($q_T^{(10)}\!<\!0.5$), the teacher's highest-confidence sampled response is
  incorrect $84\%$ of the time on code and $61\%$ on math.}
  \label{fig:motivation}
\end{figure}

The idle capacity can be used to address a more fundamental weakness of Vanilla OPD:
teacher supervision is admitted without a prompt-level reliability check. Reverse KL is
mode-seeking and concentrates the student on the teacher's high-probability behavior
\citep{gu2024minillm,thinkingmachines2025opd}. This property makes OPD efficient when the
teacher is reliable, but it can also amplify confident errors. Reliability varies by
prompt: a globally informative teacher signal may not be locally exploitable
\citep{rethinkingopd2026}, and a stronger teacher can induce negative transfer into a
smaller student \citep{smallmodels2025}. Domain routing alone does not resolve this issue
because existing multi-teacher frameworks use the selected teacher's dense reward without
verifying its reliability on the particular prompt \citep{mimo2026v2flash}.

Reliability-aware OPD has developed along two complementary directions. The first uses
distributional evidence. EOPD introduces forward KL at high-entropy teacher tokens
\citep{eopd2026}. TrOPD defines token-level trust regions from teacher--student decoding
agreement \citep{tropd2026}. REOPOLD clips and samples token rewards using likelihood ratios
and student entropy \citep{reopold2026}. These signals capture uncertainty, compatibility,
or optimization risk, but they do not directly test whether a teacher answer is correct.
Outcome evidence has also been used to regulate more local decisions. RG-OPD retains
trajectory-level distillation when verifier feedback agrees with the teacher--student
likelihood gap \citep{rgopd2026}. RLSD uses environmental correctness to determine update
direction while self-distillation modulates its magnitude \citep{rlsd2026}. At the token
level, SPOT evaluates teacher-proposed branches through verifier-scored student
continuations \citep{spot2026}. Together, these methods show how outcome feedback can
control individual trajectories or token branches. TGOPD instead uses repeated teacher
outcomes to make a prompt-level admission decision before dense supervision is applied.

Figure~\ref{fig:motivation} motivates this prompt-level decision. For the diagnostic, we
draw ten offline teacher responses per prompt and define $q_T^{(10)}(x)$ as their verifier
pass rate. The larger sample count gives a finer-grained analysis than the three fresh
online probes used during training; both quantities measure the same verifier-defined
teacher reliability. On code, self-confidence barely separates higher- and lower-reliability
prompts (AUROC~$0.51$), compared with $0.73$ on math. Within the low-reliability regime,
the teacher's highest-confidence sampled response is still incorrect $84\%$ of the time on
code and $61\%$ on math. Because reverse KL emphasizes high-probability teacher behavior,
this is precisely the type of error that Vanilla OPD can propagate.

\begin{figure}[t]
  \centering
  \includegraphics[width=\linewidth]{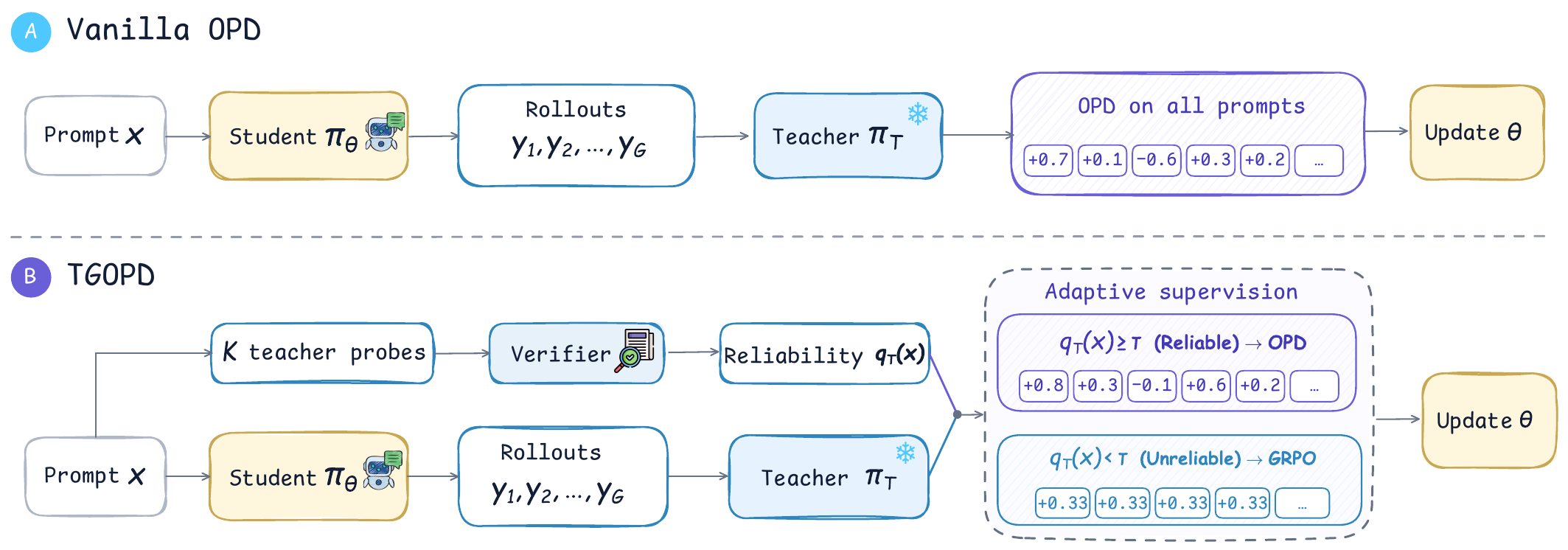}
  \caption{\textbf{TGOPD overview.}
\textbf{(A)} Vanilla OPD: the student generates $G$ on-policy rollouts, and the frozen
teacher scores every token. The resulting teacher--student log-probability gap provides
dense token-level supervision (\eqref{eq:opd-adv}) on every prompt.
\textbf{(B)} TGOPD: while the student decodes, the teacher uses its otherwise-idle capacity
to generate a small set of probe rollouts. A verifier scores these probes, and their pass
rate $q_T(x)$ (\eqref{eq:qhat}) estimates prompt-level teacher reliability. Teacher scoring
and routine verifier evaluation of student rollouts make both candidate advantages
available; the latter is omitted from the diagram for visual clarity. The gate selects
dense OPD when $q_T(x)\ge\tau$ and verifier-grounded GRPO
(\eqref{eq:grpo-adv}) otherwise. The two supervision branches are mutually exclusive.}
  \label{fig:framework}
\end{figure}

Teacher-Gated On-Policy Distillation (TGOPD) uses the teacher's idle capacity to estimate
prompt-level reliability before selecting a supervision signal. As shown in
Figure~\ref{fig:framework}, the teacher generates $K_T$ probe rollouts for the same prompt;
a verifier scores them; and their pass rate $q_T(x)=K_T^{-1}\sum_k r_k$ provides the online
reliability estimate. If $q_T(x)\!\ge\!\tau$, TGOPD uses dense OPD alone. Otherwise, it
rejects the teacher signal and uses verifier-grounded GRPO alone, provided that the student
rollout group contains reward variation. The two branches are routed rather than blended.
Because the probes run primarily while the teacher would otherwise be idle, teacher-node
GPU utilization rises from $9.8\%$ to $78.9\%$ and cluster utilization from $51.5\%$ to
$69.5\%$, with modest measured end-to-end overhead.

Across 4B and 35B students in mathematics, code, and instruction following, TGOPD
outperforms Vanilla OPD in all six domain--scale settings. Math gains most at 4B and code at 35B,
where confidence is least informative about teacher reliability. These results support the
design choice: verify the teacher at the prompt level, retain its dense signal
when the reliability check passes, and withdraw that signal when the check fails.

\section{Preliminaries}
\label{sec:prelim}

\paragraph{Setup and notation.}
Let $\pi_\theta$ denote the student policy and $\pi_T$ a frozen teacher; in a multi-teacher
setting, $\pi_T=\pi_{T,\operatorname{dom}(x)}$ is the teacher routed to the domain of prompt
$x$. For each prompt $x\sim\mathcal{D}$, the student samples a group of $G$ on-policy
rollouts $\{y^{i}\}_{i=1}^{G}\sim\pi_\theta(\cdot\mid x)$, and a binary verifier
$r(x,y)\in\{0,1\}$ scores each rollout against a ground-truth outcome: normalized answer matching for
CodeI/O output prediction, execution-based verification of predicted inputs against
reference programs for CodeI/O input prediction, and rule-based verification for
mathematics and instruction following. We write $y^{i}_{t}$ for
the $t$-th token of rollout $i$ and $\operatorname{sg}[\cdot]$ for the stop-gradient
operator.

\paragraph{On-policy distillation (OPD).}
OPD supervises every token of the student's own rollouts by minimizing the mode-seeking
reverse KL $\mathbb{KL}(\pi_\theta\,\|\,\pi_T)$. Estimated from the single sampled token at
each position, this objective takes the form of a policy-gradient update whose per-token
advantage is the teacher--student log-likelihood ratio
\citep{gu2024minillm,thinkingmachines2025opd,mimo2026v2flash}:
\begin{equation}
\hat{A}^{\mathrm{OPD}}_{i,t}
=
\beta\operatorname{sg}\!\left[
\log\pi_T\!\left(y^{i}_{t}\mid x,y^{i}_{<t}\right)
-
\log\pi_\theta\!\left(y^{i}_{t}\mid x,y^{i}_{<t}\right)
\right],
\label{eq:opd-adv}
\end{equation}
where $\beta>0$ sets the scale of the teacher signal. Every token therefore receives its
own dense, individually weighted learning signal, which contributes to OPD's sample
efficiency relative to learning from a single trajectory-level reward. Being mode-seeking,
the objective concentrates the student on the teacher's highest-probability behavior rather
than spreading probability mass over alternatives. Note that \eqref{eq:opd-adv} depends
only on $\pi_T$ and $\pi_\theta$, not on the verifier $r$: it measures how far the student
is from the teacher but does not directly assess whether the teacher's output is correct.
When $\pi_T$ is unreliable on $x$, the resulting dense gradient can therefore be
misleading.

\paragraph{GRPO with verifiable rewards.}
When teacher supervision is unavailable or withheld, the student can instead learn from
the verifier alone. GRPO \citep{deepseekmath2024} centers the reward within each group; we
omit the standard-deviation normalization \citep{liu2025drgrpo}:
\begin{equation}
\hat{A}^{\mathrm{GRPO}}_{i}
=
r\!\left(x,y^{i}\right)
-
\frac{1}{G}\sum_{j=1}^{G}r\!\left(x,y^{j}\right),
\qquad
\hat{A}^{\mathrm{GRPO}}_{i,t}
=
\hat{A}^{\mathrm{GRPO}}_{i}
\quad \forall\,t.
\label{eq:grpo-adv}
\end{equation}
Unlike the token-specific OPD advantage in \eqref{eq:opd-adv}, this signal assigns the same
scalar advantage to every token in a rollout. It is therefore trajectory-level rather than
token-specific, but remains verifier-grounded: its sign is determined by the verifier
outcome relative to the group mean.

\paragraph{The gap.}
Vanilla OPD provides dense token-level supervision without directly verifying teacher
reliability, whereas GRPO provides verifier-grounded but coarse trajectory-level
supervision. Vanilla OPD and GRPO-only training each apply a fixed supervision rule
uniformly across prompts, even though teacher reliability---and hence the preferable
learning signal---may vary from prompt to prompt. TGOPD, described next, makes this choice
per prompt by applying the same verifier to teacher probes as well as student rollouts. It
routes each prompt to exactly one signal; OPD and GRPO are never blended on the same prompt.

\section{Teacher-Gated On-Policy Distillation}
\label{sec:method}

TGOPD represents teacher reliability as a per-prompt quantity and uses it to choose between
two supervision regimes. A verifier audits the teacher on the current prompt. A successful audit selects dense OPD; otherwise, teacher supervision is withheld
and the update uses verifier-grounded GRPO. The two signals are never added or interpolated.
Figure~\ref{fig:framework} contrasts this procedure with vanilla OPD. In panel~A, the
teacher's log-probabilities enter every update. In panel~B, they enter only after the audit
passes; a verifier-grounded signal is used when the audit fails.
Section~\ref{sec:method:gate} defines the reliability estimate and the gate it drives,
\S\ref{sec:method:objective} gives the objective the gate induces, and
\S\ref{sec:method:algo} shows how the audit overlaps with the asynchronous training loop's
otherwise-idle teacher window.

\subsection{Teacher Reliability Probe and Gate}
\label{sec:method:gate}

The gate requires an estimate of teacher reliability for each prompt, which the vanilla OPD loop does not provide. We define this quantity below and estimate it with a small
number of teacher samples.

\paragraph{Reliability and its estimator.}
We define the teacher's reliability on a prompt as its expected verifier reward,
\begin{equation}
R_T(x) \;=\; \mathbb{E}_{y\sim\pi_T(\cdot\mid x)}\big[r(x,y)\big]\;\in\;[0,1],
\label{eq:reliability}
\end{equation}
that is, the probability that a sample from the teacher solves $x$.  Since $R_T$ is not
available in closed form, we estimate it with a small teacher probe: the teacher
independently generates $K_T$ rollouts
$\{\hat{y}^{k}\}_{k=1}^{K_T}\sim\pi_T(\cdot\mid x)$ on the same prompt. The verifier used
for student rollouts scores each teacher rollout, and their empirical pass rate is
(Figure~\ref{fig:framework}B)
\begin{equation}
q_T(x)\;=\;\frac{1}{K_T}\sum_{k=1}^{K_T} r\!\left(x,\hat{y}^{k}\right).
\label{eq:qhat}
\end{equation}
Because the probes are i.i.d.\ draws from $\pi_T(\cdot\mid x)$ and $r$ is binary,
$K_T\,q_T(x)\sim\mathrm{Binomial}\!\left(K_T,R_T(x)\right)$.  The estimator is therefore
unbiased, $\mathbb{E}[q_T(x)]=R_T(x)$, for any probe budget, with variance
$R_T(x)\big(1-R_T(x)\big)/K_T$.  Unbiasedness alone does not imply that a particular
finite probe budget yields reliable accept/reject decisions.  We empirically find
$K_T=3$ sufficient in our main experiments; Section~\ref{ssec:threshold} uses a larger
budget to examine the threshold at a finer resolution.

\paragraph{Verifier-grounded reliability signal.}
The estimator $q_T(x)$ uses task outcomes rather than the teacher's own
confidence. Entropy and teacher--student likelihood agreement are functions of model
distributions alone: they can quantify uncertainty or compatibility, but cannot by
themselves distinguish equally confident correct and incorrect answers. Evaluating $r$ on
completed teacher outputs provides direct outcome evidence. Unlike verifier-aware token or
trajectory gates, $q_T(x)$ aggregates repeated teacher outcomes into a prompt-level
reliability estimate before the supervision branch is selected.

\paragraph{The reliability gate.}
A hard gate admits the teacher only when its estimated reliability reaches a threshold
$\tau\in(0,1]$ (Figure~\ref{fig:framework}B, ``reliability gate''):
\begin{equation}
g(x)\;=\;\mathbb{1}\!\left[\,q_T(x)\;\ge\;\tau\,\right]\;\in\;\{0,1\}.
\label{eq:gate}
\end{equation}
Since $q_T(x)$ takes values in $\{0,1/K_T,\dots,1\}$, the gate opens exactly when at least
$\lceil\tau K_T\rceil$ of the $K_T$ probes pass; only the pair
$\big(\lceil\tau K_T\rceil,\,K_T\big)$, and not $\tau$ in isolation, is operationally
meaningful.  We use $K_T=3$ and $\tau=2/3$ in all main experiments, i.e.\ a two-of-three majority.
Section~\ref{ssec:threshold} sweeps the threshold at $K_T=5$ and finds a broad optimum
around a simple majority. This result suggests that a coarse reliability decision is
sufficient; the gate need not rank prompts precisely.

\subsection{Gate-Conditioned Supervision Routing}
\label{sec:method:objective}

For each prompt, the gate selects one of the two supervision signals defined in
\S\ref{sec:prelim}. When it is open, the dense OPD signal controls the update and the
trajectory-level verifier reward does not enter the gradient. When the gate is closed, teacher supervision is withheld and the
verifier-grounded GRPO signal controls the update. Withholding teacher supervision
therefore does not necessarily discard the training prompt. Whenever
the student's rollout group contains reward variation, the group-relative fallback assigns
positive advantage to above-average attempts and negative advantage to below-average ones
(Figure~\ref{fig:framework}B, lower signal box). If every rollout receives the same reward,
however, the centered GRPO advantage is exactly zero and that prompt produces no update.
For pipeline regularity, both candidate advantages are formed for every prompt before the
update. This eager computation does not imply additive supervision: the binary gate selects
exactly one candidate, giving the per-token advantage
\begin{equation}
\hat{A}^{\mathrm{TGOPD}}_{i,t}
\;=\; g(x)\,\hat{A}^{\mathrm{OPD}}_{i,t}
\;+\;\big(1-g(x)\big)\,\hat{A}^{\mathrm{GRPO}}_{i,t}.
\label{eq:tgopd-adv}
\end{equation}
Because $g(x)\in\{0,1\}$, \eqref{eq:tgopd-adv} is a selector rather than an
interpolation: OPD and GRPO never jointly supervise the same prompt. Vanilla OPD and pure
GRPO are its two degenerate endpoints, obtained by fixing $g(x)=1$ and $g(x)=0$,
respectively, for every prompt.

The resulting advantage enters the standard clipped policy-gradient surrogate. Let
$\pi_{\mathrm{old}}$ denote the frozen rollout-policy snapshot and define
$\rho_{i,t}(\theta)=\pi_\theta(y^i_t\mid x,y^i_{<t})/
\pi_{\mathrm{old}}(y^i_t\mid x,y^i_{<t})$. We optimize
\begin{equation}
\mathcal{L}_{\mathrm{TGOPD}}(\theta)
=-\,\mathbb{E}_{\substack{x\sim\mathcal{D}\\
                   \{y^i\}_{i=1}^{G}\sim\pi_{\mathrm{old}}(\cdot\mid x)}}
\left[\frac{1}{\sum_i |y^{i}|}
\sum_{i=1}^{G}\sum_{t=1}^{|y^{i}|}
\min\!\left(
\rho_{i,t}(\theta)\hat{A}^{\mathrm{TGOPD}}_{i,t},
\operatorname{clip}\!\left(\rho_{i,t}(\theta),1-\epsilon,1+\epsilon\right)
\hat{A}^{\mathrm{TGOPD}}_{i,t}
\right)\right].
\label{eq:tgopd-loss}
\end{equation}
The advantage is held fixed within the update. The asynchronous implementation additionally
uses IcePop to suppress excessive train--inference mismatch; this system-level correction
does not alter the supervision-routing objective above (\S\ref{ssec:setup}).

\paragraph{Un-normalized fallback.}
Our implementation omits the standard-deviation normalization of GRPO
\citep{liu2025drgrpo}. This avoids rescaling the fallback by the within-group reward
standard deviation. We use it as an implementation convention, not as a requirement of the
gate. A controlled comparison with normalized variants is left to future work.

\paragraph{Design choices.}
The gate acts at the prompt level: a prompt is supervised either entirely by
OPD or entirely by the verifier-grounded fallback, never by a per-token or additive mixture.
This avoids entangling two signals of different density and provenance within a single
trajectory, where the effective supervision would depend on an arbitrary interpolation
weight. The gate also withdraws teacher supervision only where the audit fails, so on the
majority of prompts the full direction and magnitude of the dense OPD signal survives
intact. Section~\ref{ssec:main-results} shows the cost of reducing the teacher's role on
every prompt rather than only on those that fail the audit.

\subsection{Algorithm and System Realization}
\label{sec:method:algo}

Algorithm~\ref{alg:tgopd} summarizes one rollout-update cycle. Relative to vanilla OPD,
TGOPD retains the same student rollout and teacher scoring operations and adds the probe in
\eqref{eq:qhat}, which determines the gate in \eqref{eq:gate}. Teacher probes run concurrently
with student generation, using otherwise-idle teacher capacity; the measured residual
overhead is summarized in Appendix~\ref{app:gpu-traces}.

\begin{algorithm}[H]
\caption{TGOPD: one asynchronous rollout--update cycle}
\label{alg:tgopd}
\begin{algorithmic}[1]
\Require current student $\pi_\theta$; frozen rollout snapshot $\pi_{\mathrm{old}}$;
frozen teacher $\pi_T$; verifier $r$
\Require prompt batch $\mathcal{B}$; group size $G$; probe budget $K_T$; threshold $\tau$;
PPO clip $\epsilon$; step size $\eta$
\Ensure updated student parameters $\theta$
\For{each prompt $x\in\mathcal{B}$}
  \State \textbf{concurrently:} \Comment{probe fills the teacher's idle window}
  \State \quad \emph{rollout nodes:}\;
         $\{y^{i}\}_{i=1}^{G}\sim\pi_{\mathrm{old}}(\cdot\mid x)$; record rollout log-probs
         \Comment{nominal critical path}
  \State \quad \emph{teacher node:}\;
         $\{\hat{y}^{k}\}_{k=1}^{K_T}\sim\pi_T(\cdot\mid x)$
  \State $q_T(x)\gets\frac{1}{K_T}\sum_{k=1}^{K_T} r\!\left(x,\hat{y}^{k}\right)$
         \Comment{reliability estimate, \eqref{eq:qhat}}
  \Statex \hspace{\algorithmicindent}\emph{// both candidate signals are formed for every prompt}
  \State evaluate $\log\pi_T\!\left(y^{i}_{t}\mid x,y^{i}_{<t}\right)$ for all $i,t$
         \Comment{teacher scoring pass}
  \State score the student rollouts: $r\!\left(x,y^{i}\right)$ for $i=1,\dots,G$
         \Comment{the RLVR signal}
  \If{$q_T(x)\ge\tau$} \Comment{teacher supervision admitted}
     \State $\hat{A}_{i,t}\gets\hat{A}^{\mathrm{OPD}}_{i,t}$ for all $i,t$
            \Comment{\eqref{eq:opd-adv}}
  \Else \Comment{teacher supervision withheld}
     \State $\hat{A}_{i,t}\gets\hat{A}^{\mathrm{GRPO}}_{i,t}$ for all $i,t$
            \Comment{\eqref{eq:grpo-adv}}
  \EndIf
\EndFor
\State form $\rho_{i,t}(\theta)$ from current and recorded rollout log-probs
\State $\theta\gets\theta-\eta\,\nabla_\theta\mathcal{L}_{\mathrm{TGOPD}}(\theta)$
       over $\mathcal{B}$ \Comment{\eqref{eq:tgopd-loss}}
\end{algorithmic}
\end{algorithm}

\paragraph{Overlapping probes with teacher idle time.}
In an asynchronous deployment, the teacher's standard task is a forward pass over
already-generated tokens. This is far cheaper than the student's autoregressive decoding and
cannot begin until a batch of student rollouts is ready, so the teacher node idles for most of each cycle
(\S\ref{ssec:gpu-util}). TGOPD issues the $K_T$ probe rollouts at the start of the
cycle, concurrently with student generation, so most probe work occupies this window. If
all $K_T$ decodes finish before the student-side batch completes, the probe adds no
wall-clock cost; in practice the overlap is substantial but not perfect
(Appendix~\ref{app:gpu-traces}).

\paragraph{Scoring stays unconditional.}
Teacher scoring runs on every prompt, exactly as under vanilla OPD, and the verifier scores
the student's rollouts because that reward is the reinforcement-learning signal in any case.
The scoring pass is a single forward evaluation over tokens the student has already
produced. It is batched as rollouts arrive and runs on a node that is not the throughput
bottleneck. Making this pass conditional for a minority of prompts would add a
data-dependent pipeline branch while saving little computation. The additional work in
TGOPD therefore consists of the $K_T$ probe decodes,
which is why teacher utilization can rise from $9.8\%$ to $78.9\%$ with only modest
end-to-end overhead (\S\ref{ssec:gpu-util}): most audit work consumes capacity that was
previously wasted, while the measured remainder slightly extends the cycle.

\section{Experiments}
\label{sec:experiments}
\definecolor{tgopdRowTint}{RGB}{235,243,250}
\definecolor{tgopdGroupTint}{RGB}{243,244,246}

\subsection{Experimental Setup}
\label{ssec:setup}

\paragraph{Models and teachers.}
We evaluate at two scales: a dense 4B student (Qwen3.5-4B) and a mixture-of-experts 35B
student (Qwen3.6-35B-A3B, with 3B active parameters).  For each of three
domains (mathematics, code, and instruction following, or IF), we train a dedicated
domain-specialist teacher by applying GRPO~\citep{deepseekmath2024} to the same base model.
Teachers are frozen after training and used to generate reliability probes and score
student rollouts; their own evaluation
scores provide both a teacher-performance reference and a GRPO-only reference on the same
architecture (the ``Teacher'' row in Table~\ref{tab:main}).
All runs use the slime framework~\citep{slime_github} with an asynchronous
rollout--update loop that overlaps teacher scoring with student generation.
Student updates use the standard PPO-clipped surrogate in \eqref{eq:tgopd-loss}. The
implementation additionally applies IcePop~\citep{ring1t2025} to stabilize the
train--inference probability mismatch introduced by asynchronous rollout; this correction is
shared by all compared methods and is not part of the TGOPD gate.
Within TGOPD, the OPD and GRPO advantages are mutually exclusive: the former is used only
when the teacher passes the prompt-level audit, and the latter only when it fails
(\eqref{eq:tgopd-adv}).
Appendix~\ref{app:training-detail} summarizes the recorded optimization and infrastructure
settings.

\paragraph{Baselines.}
All comparisons share the same student initialization, frozen domain teacher, training corpus,
and training pipeline. Vanilla OPD~\citep{thinkingmachines2025opd} applies the sampled-token
reverse-KL advantage in \eqref{eq:opd-adv} to every prompt. TrOPD~\citep{tropd2026} uses an
adaptive token-level trust region, applying reverse KL within the region, forward KL to
outliers, and teacher-prefix off-policy guidance. RG-OPD~\citep{rgopd2026} retains
trajectory-level distillation only when the verifier-derived advantage agrees with the
teacher--student likelihood gap; inconsistent trajectories are omitted from its
distillation objective. We also include an RLSD-style baseline inspired by
RLSD~\citep{rlsd2026}: the verifier-derived advantage determines update direction, while
the same external frozen teacher used by the other baselines rescales token-level magnitude.
This variant adopts RLSD's direction--magnitude decomposition without reproducing its
privileged-context self-distillation setting.

\paragraph{Training data and evaluation.}
Mathematics prompts are drawn from DAPO-Math-17K~\citep{yu2026dapo}; code prompts are the
input--output prediction questions of CodeI/O~\citep{li2025codeio}; and IF prompts are
filtered from Nemotron-Cascade~2~\citep{yang2026nemotron}.  Each method is trained per
domain from the same student initialization and the same prompt pool.
We report in-domain benchmarks: AIME~2025 and AIME~2026 (accuracy, 64-run average),
HMMT-Feb 2025 (accuracy, 32-run average) for mathematics; the code-generation subtask of
LiveCodeBench (pass@1, 6-run average) and OJBench overall (C++/Python aggregate) for code;
and IFBench and IFEval for instruction following.
Appendix~\ref{app:eval-detail} summarizes the evaluation protocols and the number of
independent evaluation generations for each benchmark.

\subsection{Main Results}
\label{ssec:main-results}

\begin{table}[!t]
\centering
\caption{In-domain distillation across two model scales and three domains.
Each domain uses a separately trained student and its corresponding
GRPO teacher.
All scores are higher-is-better.
Avg.\ is the unweighted mean of the seven benchmark scores,
aggregated across the three domain-specific students.
\textbf{Bold} marks the best score among distillation methods within
each model scale, including ties at the displayed precision;
\underline{underlining} marks benchmark scores above the corresponding
domain teacher.}
\label{tab:main}

\begingroup
\setlength{\tabcolsep}{4pt}
\renewcommand{\arraystretch}{1.08}

\resizebox{\linewidth}{!}{%
\begin{tabular}{@{}l @{\hskip 4pt} ccc
@{\hskip 7pt}@{\hskip 7pt} cc
@{\hskip 7pt}@{\hskip 7pt} cc
@{\hskip 7pt}@{\hskip 7pt} c@{}}
\toprule
& \multicolumn{3}{c}{\textsc{Math}}
& \multicolumn{2}{c}{\textsc{Code}}
& \multicolumn{2}{c}{\textsc{IF}}
& \\
\cmidrule(lr){2-4}
\cmidrule(lr){5-6}
\cmidrule(lr){7-8}
Method
& AIME 2025 & AIME 2026 & HMMT-Feb
& LiveCodeBench & OJBench
& IFBench & IFEval
& Avg. \\
\midrule

\rowcolor{tgopdGroupTint}
\multicolumn{9}{@{}l}{\textit{Qwen3.5-4B}} \\
\addlinespace[2pt]
Base Model
& 47.8 & 58.1 & 40.5
& 39.4 & 14.4
& 35.9 & 85.3
& 45.91 \\
Teacher
& 63.4 & 73.4 & 54.8
& 53.3 & 17.0
& 55.9 & 85.3
& 57.59 \\
\cdashline{1-9}[2pt/3pt]
\noalign{\vskip 2pt}
Vanilla OPD
& 61.1 & 71.2 & 54.3
& 45.0 & \underline{18.1}
& 47.8 & 84.5
& 54.57 \\
TrOPD
& 62.6 & 70.4 & \underline{\textbf{57.7}}
& \textbf{47.8} & \underline{19.4}
& 53.3 & 84.4
& 56.51 \\
RG-OPD
& \underline{64.6} & 72.7 & 54.2
& 45.6 & \underline{19.4}
& \textbf{53.5} & 84.0
& 56.29 \\
RLSD-style
& 46.8 & 57.0 & 42.4
& 38.8 & 13.8
& 33.4 & 79.8
& 44.57 \\
\addlinespace[2pt]
\rowcolor{tgopdRowTint}
\textbf{TGOPD (ours)}
& \underline{\textbf{64.8}}
& \underline{\textbf{73.5}}
& \underline{55.7}
& 47.1 & \underline{\textbf{20.0}}
& 50.4 & \textbf{85.2}
& \textbf{56.67} \\

\midrule
\rowcolor{tgopdGroupTint}
\multicolumn{9}{@{}l}{\textit{Qwen3.6-35B-A3B}} \\
\addlinespace[2pt]
Base Model
& 69.8 & 74.4 & 64.3
& 60.0 & 25.9
& 30.0 & 86.8
& 58.74 \\
Teacher
& 76.3 & 79.8 & 65.3
& 62.7 & 27.6
& 52.1 & 90.8
& 64.94 \\
\cdashline{1-9}[2pt/3pt]
\noalign{\vskip 2pt}
Vanilla OPD
& 73.4 & 77.3 & \textbf{64.1}
& 60.2 & \underline{28.0}
& 44.2 & 89.9
& 62.44 \\
TrOPD
& \textbf{75.0} & 77.9 & 63.9
& 57.8 & \underline{\textbf{28.7}}
& \textbf{44.5} & 90.8
& 62.66 \\
RG-OPD
& 74.8 & 78.1 & 62.8
& 57.0 & 26.9
& 44.2 & 90.6
& 62.06 \\
RLSD-style
& 72.7 & 75.1 & 60.3
& 57.7 & 26.5
& 36.4 & 85.7
& 59.20 \\
\addlinespace[2pt]
\rowcolor{tgopdRowTint}
\textbf{TGOPD (ours)}
& 74.0 & \textbf{78.7} & \textbf{64.1}
& \underline{\textbf{64.0}}
& \underline{\textbf{28.7}}
& 43.7 & \underline{\textbf{91.5}}
& \textbf{63.53} \\
\bottomrule
\end{tabular}%
}
\endgroup

\end{table}

Table~\ref{tab:main} reports in-domain distillation results across
two model scales, with a separate student trained for each domain.
TGOPD achieves the highest seven-benchmark average among the
compared distillation methods at both scales:
$56.67$ at 4B and $63.53$ at 35B.
These scores exceed Vanilla OPD by $2.10$ and $1.09$ points,
respectively, and the strongest baseline by average, TrOPD,
by $0.16$ and $0.87$ points.
Across the 14 benchmark--scale pairs, TGOPD achieves the best
distillation score on nine, including ties, and surpasses the
corresponding domain teacher on seven.
The overall gains coexist with benchmark-specific differences:
TrOPD remains strongest on 4B HMMT-Feb and LiveCodeBench,
while TGOPD ties TrOPD on 35B OJBench and Vanilla OPD on
35B HMMT-Feb.

The gains over Vanilla OPD extend to the within-domain average
in all six domain--scale settings.
At 4B, the improvements are $2.47$ points in math, $2.00$ in code,
and $1.65$ in instruction following.
At 35B, the corresponding gains are $0.67$, $2.25$, and $0.55$ points.
Math gains most at 4B and code at 35B.
The code gains are consistent with the diagnostic in
Figure~\ref{fig:motivation}, where teacher confidence provides
little separation between high- and low-reliability code prompts
(AUROC~$0.51$), motivating reliability checks based on verified
teacher outcomes.
The domain averages also mask variation across benchmarks:
on 35B IFBench, TGOPD scores $43.7$, slightly below Vanilla OPD's
$44.2$, while improving IFEval from $89.9$ to $91.5$.

The 35B LiveCodeBench results provide the clearest example of
the benefit of selective teacher supervision.
Vanilla OPD scores $60.2$, improving only $0.2$ points over the base
model's $60.0$.
TrOPD, RG-OPD, and RLSD-style fall below the base model,
by $2.2$, $3.0$, and $2.3$ points, respectively.
TGOPD reaches $64.0$, improving over the base by $4.0$ points
and over Vanilla OPD by $3.8$ points.
It also exceeds the code teacher by $1.3$ points on LiveCodeBench
and $1.1$ points on OJBench, where it ties TrOPD at $28.7$.
At 4B, TGOPD improves LiveCodeBench from Vanilla OPD's $45.0$
to $47.1$, closing approximately $55\%$ of the base-to-teacher
gap, compared with $40\%$ for Vanilla OPD.
Together, these results support prompt-level selection of
teacher supervision in code, where confidence alone is a weak
indicator of reliability.

The RLSD-style variant provides a complementary comparison
of how teacher information is used.
It uses the teacher to scale update magnitude and verifier
feedback to determine update direction, applying this rule
to every prompt.
The variant falls below the base model on nine of the
14 benchmark--scale pairs, including a $5.5$-point drop on
4B IFEval and a $2.3$-point drop on 35B LiveCodeBench.
TGOPD instead retains the full OPD signal on prompts that pass
the reliability check and removes that signal on prompts
that fail.
This comparison is consistent with the value of retaining
dense teacher guidance selectively.
The following ablations more directly examine the contributions
of teacher-signal filtering and the GRPO fallback.

\subsection{Extension to Multi-Domain OPD}
\label{ssec:mopd}

\begin{table}[!t]
\centering
\caption{TGOPD applied to multi-domain OPD (MOPD).
Each run trains a single student on math, code, and IF simultaneously, routing
prompts to domain-specialist teachers.
Results are reported at 199 training steps with $K_T\!=\!3,\;\tau\!=\!2/3$.
\textbf{Bold}: best per column at the displayed precision, including ties.
Avg.\ is computed before display rounding.}
\label{tab:mopd}
\small
\setlength{\tabcolsep}{4pt}
\renewcommand{\arraystretch}{1.12}
\resizebox{\linewidth}{!}{%
\begin{tabular}{@{}l c
  @{\hskip 10pt} ccc
  @{\hskip 10pt} cc
  @{\hskip 10pt} cc@{}}
\toprule
& & \multicolumn{3}{c}{\textsc{Math}}
& \multicolumn{2}{c}{\textsc{Code}}
& \multicolumn{2}{c}{\textsc{IF}} \\
\cmidrule(lr){3-5}\cmidrule(lr){6-7}\cmidrule(lr){8-9}
Method & Avg.
  & AIME 25 & AIME 26 & HMMT-Feb
  & LCB & OJBench
  & IFBench & IFEval \\
\midrule
\multicolumn{9}{@{}>{\columncolor{tgopdGroupTint}[0pt][0pt]}l@{}}{\emph{Qwen3.5-4B}} \\[2pt]
MOPD
  & 53.40
  & \textbf{58.1} & 63.8 & 50.7
  & \textbf{50.5} & 18.1
  & 48.8 & 83.9 \\
MOPD + TGOPD
  & \textbf{54.54}
  & \textbf{58.1} & \textbf{67.6} & \textbf{51.3}
  & 50.0 & \textbf{19.0}
  & \textbf{51.7} & \textbf{84.1} \\
\midrule
\multicolumn{9}{@{}>{\columncolor{tgopdGroupTint}[0pt][0pt]}l@{}}{\emph{Qwen3.6-35B-A3B}} \\[2pt]
MOPD
  & 60.99
  & \textbf{73.1} & 77.3 & 60.7
  & 57.4 & 23.3
  & \textbf{44.4} & 90.7 \\
MOPD + TGOPD
  & \textbf{61.94}
  & 73.0 & \textbf{78.0} & \textbf{62.4}
  & \textbf{57.8} & \textbf{27.6}
  & 44.0 & \textbf{90.8} \\
\bottomrule
\end{tabular}
}
\end{table}

The experiments above evaluate TGOPD in the single-domain setting, where each run trains
one student with one domain teacher. We also consider multi-domain OPD
(MOPD) \citep{mimo2026v2flash,mopd2026}, in which a single student is trained on math,
code, and IF prompts simultaneously, with each prompt routed to a domain-specialist teacher.
Because the gating decision is per-prompt and per-teacher, TGOPD applies to MOPD without
modification: the routed teacher generates $K_T\!=\!3$ probes on its assigned prompt, the domain verifier scores them, and the gate admits or withholds teacher
supervision as before.

Table~\ref{tab:mopd} reports results at 199 training steps. At the 4B scale, TGOPD
improves the seven-benchmark average from $53.40$ to $54.54$ ($+1.14$), improving 6 of 7
benchmark scores before display rounding; at the 35B scale the average rises from $60.99$ to $61.94$ ($+0.95$), improving
5 of 7 benchmark scores before display \mbox{rounding}. \mbox{The largest} single-benchmark gains appear on AIME~2026 ($+3.85$ at 4B),
IFBench ($+2.94$ at 4B), and OJBench ($+0.87$ at 4B, $+4.31$ at 35B); the code result in
particular is consistent with the single-domain finding, since confidence is least
informative about teacher reliability on code. The few regressions (LCB at 4B,
$-0.48$; AIME~2025, $-0.15$; IFBench, $-0.41$, at 35B) are all under half a point.
The average improvements at both scales show that the same per-prompt gate can be used in
multi-domain distillation without reducing overall quality.

\subsection{Reclaiming Idle Teacher Capacity}
\label{ssec:gpu-util}

\begin{figure}[!t]
\centering
\includegraphics[width=\linewidth]{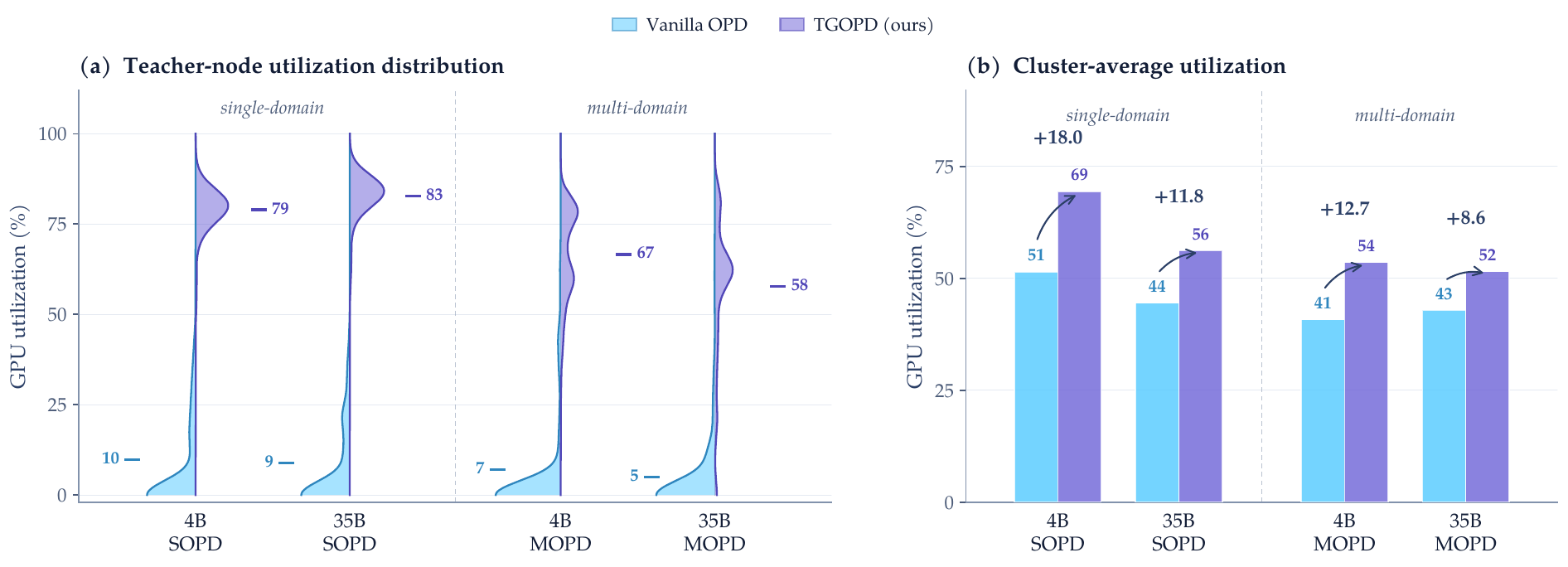}
\caption{\textbf{GPU utilization before and after TGOPD, across four configurations.}
\textbf{(a)} Split-violin view of teacher-node utilization.
Under vanilla OPD the mass piles up near $0\%$ in every configuration; TGOPD shifts the
whole distribution upward, showing that probes use the idle window.
The MOPD distributions under TGOPD are visibly bimodal because probing is domain-routed:
the teacher node is partitioned into three domain-specialist engines. Each prompt activates
only its corresponding teacher, while different prompts can use different teachers
concurrently, allowing node occupancy to vary with the batch's domain mix.
\textbf{(b)} Cluster-average utilization. Gains track both the teacher's share of the
cluster ($1/5$ at 4B vs.\ $1/7$ at 35B) and the saturation the probe achieves
(higher under SOPD than under MOPD).}
\label{fig:gpu-util}
\end{figure}

\suppressfloats[t]
\begin{table}[!htbp]
\centering
\caption{GPU utilization under Vanilla OPD and TGOPD across four configurations:
single-domain OPD (SOPD) and multi-domain OPD (MOPD) at both model scales.
Each measurement covers a one-hour window; \emph{idle} denotes the fraction
of 15-second samples with utilization below $5\%$.
Student-rollout utilization remains similar between the two methods,
with differences of at most $2.1$ percentage points.
In MOPD, one teacher node hosts three frozen domain-specialist engines,
allocated 4 GPUs for code, 2 for math, and 2 for IF.
Reported rollout utilization is averaged over the two student-rollout nodes.}
\label{tab:gpu-util}
\small
\setlength{\tabcolsep}{4pt}
\renewcommand{\arraystretch}{1.12}
\begin{tabular}{@{}p{\dimexpr0.18\linewidth-8pt\relax}*{8}{>{\centering\arraybackslash}p{\dimexpr0.1025\linewidth-7pt\relax}}@{}}
\toprule
& \multicolumn{2}{c}{Teacher utilization}
& \multicolumn{2}{c}{Teacher idle}
& \multicolumn{2}{c}{Rollout nodes}
& \multicolumn{2}{c}{Cluster average} \\
\cmidrule(lr){2-3}\cmidrule(lr){4-5}\cmidrule(lr){6-7}\cmidrule(lr){8-9}
Configuration & OPD & TGOPD & OPD & TGOPD & OPD & TGOPD & OPD & TGOPD \\
\midrule
\multicolumn{9}{@{}>{\columncolor{tgopdGroupTint}[0pt][0pt]}l@{}}{\emph{Qwen3.5-4B}\quad (5 nodes, teacher = $1/5$)} \\[2pt]
SOPD & 9.8\% & \textbf{78.9\%} & 59.0\% & 0.0\% & 69.6\% & 70.5\% & 51.5\% & \textbf{69.5\%} \\
MOPD & 7.0\% & \textbf{66.6\%} & 78.0\% & 0.0\% & 64.2\% & 66.3\% & 40.8\% & \textbf{53.5\%} \\
\midrule
\multicolumn{9}{@{}>{\columncolor{tgopdGroupTint}[0pt][0pt]}l@{}}{\emph{Qwen3.6-35B-A3B}\quad (7 nodes, teacher = $1/7$)} \\[2pt]
SOPD & 8.8\% & \textbf{82.8\%} & 57.0\% & 0.0\% & 73.6\% & 73.8\% & 44.5\% & \textbf{56.3\%} \\
MOPD & 5.0\% & \textbf{57.7\%} & 70.0\% & 2.0\% & 65.6\% & 63.8\% & 42.9\% & \textbf{51.5\%} \\
\bottomrule
\end{tabular}
\end{table}

Figure~\ref{fig:idle} showed low teacher utilization at the 4B scale. We extend the analysis
to both model scales and to single- and multi-domain OPD, giving four configurations in
total. We also examine whether probe generation affects other workloads.
Table~\ref{tab:gpu-util} and Figure~\ref{fig:gpu-util} report
summary statistics; the underlying one-hour utilization traces for all four configurations
are plotted in Appendix~\ref{app:gpu-traces}.

\paragraph{The idle pattern is structural, not configuration-specific.}
Table~\ref{tab:gpu-util} and Figure~\ref{fig:gpu-util} show the same profile in every
configuration. Without probes, the teacher node averages between $5.0\%$ and $9.8\%$
utilization, with $57$--$78\%$ of all 15-second samples falling below $5\%$.
The cause is architectural rather than incidental. The teacher's scoring pass is a single
forward evaluation over already-generated tokens, so it completes in a small fraction of the
time the student spends on autoregressive decoding, and it cannot begin until the student
finishes. The teacher therefore waits by construction, at every scale and in both training
modes.
Multi-domain OPD is the more wasteful of the two modes: its teacher node idles for
$78\%$ (4B) and $70\%$ (35B) of samples, against $57$--$59\%$ under SOPD. This follows from
how the node is partitioned: three domain-specialist engines share it, and each runs only
on prompts routed to its own domain. Any individual engine, and hence the node average,
sits at zero for a larger share of the run.

\paragraph{Higher teacher utilization with limited measured runtime overhead.}
Adding teacher probes raises mean teacher-node utilization from
$9.8\%$ to $78.9\%$ and from $8.8\%$ to $82.8\%$ under SOPD
at 4B and 35B, respectively.
Under MOPD, utilization rises from $7.0\%$ to $66.6\%$ at 4B
and from $5.0\%$ to $57.7\%$ at 35B.
The fraction of samples below $5\%$ utilization falls to $0$--$2\%$
across the four configurations.
Student-rollout utilization changes by between $-1.8$ and $+2.1$
percentage points.
These measurements show that probing makes greater use of the allocated
teacher capacity, but utilization alone does not determine resource
contention or end-to-end training efficiency.
In the matched 35B CodeIO comparison, mean step time increases by $5.9\%$
over $500$ aligned rollout cycles
(Appendix~\ref{app:gpu-traces}).

\paragraph{Cluster gains track teacher fraction and probe saturation.}
Cluster-wide utilization improves by $+18.0$ and $+11.8$ points under SOPD and by $+12.7$
and $+8.6$ under MOPD (4B, 35B respectively). The ordering follows from two factors. At 4B,
the teacher accounts for $1/5$ of the nodes, compared with $1/7$ at 35B, so the same
teacher-side gain has a larger effect on the cluster average. Probe saturation is also
lower under MOPD because the teacher node is partitioned by domain. Decomposing the cluster delta into
per-role contributions shows that the teacher term dominates in every case: it accounts
for $13.8$ of the $+18.0$ at 4B~SOPD, $10.6$ of $+11.8$ at 35B~SOPD, $11.9$ of $+12.7$ at
4B~MOPD, and $7.5$ of $+8.6$ at 35B~MOPD. The residual in each case comes from ordinary
run-to-run drift on the train nodes and is largest at 4B~SOPD ($+3.8$); in the two MOPD
runs the train nodes move by $-0.0$ and $+1.6$ points, so essentially the entire gain is
reclaimed teacher capacity.

\paragraph{Consistency across deployment topologies.}
The same pattern appears in single- and multi-domain OPD despite their different teacher
allocations. SOPD dedicates a full node to one teacher, whereas MOPD partitions that node
across three domain specialists that are individually idle more often. In both cases, most
probe work uses previously idle capacity. The utilization gain therefore does not depend on
one particular teacher layout.

\subsection{Ablation: Gate Threshold}
\label{ssec:threshold}

\begin{wrapfigure}[18]{r}{0.45\linewidth}
  \centering
  \vspace{-8pt}
  \includegraphics[width=\linewidth]{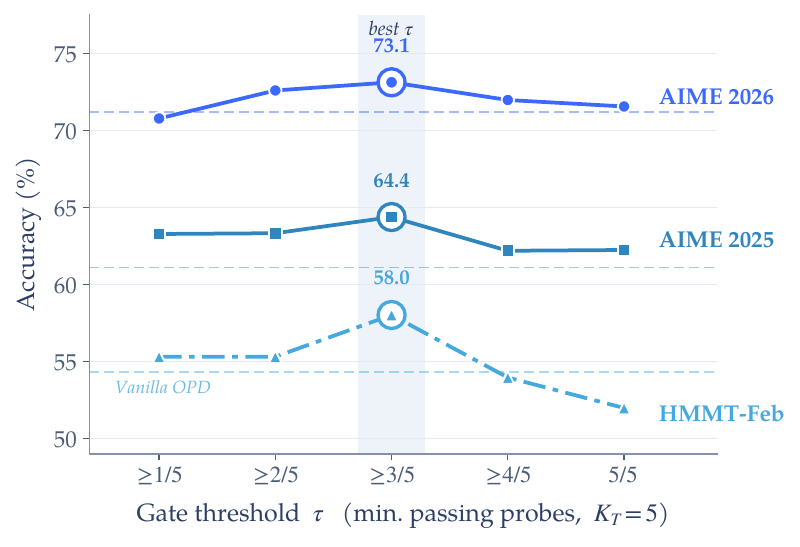}
  \caption{\textbf{Gate-threshold sweep on math benchmarks}
  (Qwen3.5-4B, $K_T\!=\!5$, 99 steps). Solid: TGOPD; dotted:
  Vanilla OPD. All three benchmarks peak near $\tau\!=\!3/5$.}
  \label{fig:threshold}
\end{wrapfigure}

The main experiments fix $\tau\!=\!2/3$ and $K_T\!=\!3$.
For this diagnostic ablation, we increase the probe budget to $K_T\!=\!5$, yielding five
nontrivial pass-count cutoffs and thus a finer-grained view of threshold sensitivity.
We sweep $\tau$ across $\{1/5,\;2/5,\;3/5,\;4/5,\;5/5\}$ while keeping all other
hyperparameters identical (Qwen3.5-4B, math domain, 99 training steps).

Figure~\ref{fig:threshold} shows an inverted-U relationship between gate strictness and
downstream accuracy. At low thresholds ($\tau\!\leq\!2/5$), the pass requirement
admits many lower-reliability teacher signals, including confidently wrong ones. Performance still
exceeds Vanilla OPD on AIME~2025 ($+2.2$) and HMMT-Feb ($+1.0$),
but the gains are smaller because the gate provides little filtering. At
$\tau\!=\!3/5$, all three benchmarks reach their
peak: AIME~2026 scores $73.1$, AIME~2025 scores $64.4$, and HMMT-Feb scores $58.0$.
These values exceed Vanilla OPD by $+1.9$, $+3.3$, and $+3.7$, respectively.
At high thresholds ($\tau\!\geq\!4/5$), many useful teacher signals are rejected.
HMMT-Feb falls below the Vanilla OPD baseline at $\tau\!=\!5/5$
($52.0$ vs.\ $54.3$), which is consistent with useful OPD signal being discarded.

The peak at $\tau=3/5$ (pass rate $60\%$) is close to the default
$\tau=2/3$ ($\approx67\%$) used in the main experiments. Under the
$K_T=5$ diagnostic setting, the results indicate a broad performance
optimum around a majority-vote criterion.

\WFclear
\newsavebox{\tgopdFallbackTable}
\begin{lrbox}{\tgopdFallbackTable}
\begin{minipage}{0.56\linewidth}
\makeatletter\def\@captype{table}\makeatother
  \centering
  \caption{Closed-gate policy on single-domain mathematics runs.
  Vanilla OPD and GRPO fallback use the corresponding math results in
  Table~\ref{tab:main}. Avg.\ is the mean of the three benchmark scores
  before display rounding. \textbf{Bold}: best within each model scale
  at the displayed precision, including ties.}
  \label{tab:fallback-sopd}
  \footnotesize
  \setlength{\tabcolsep}{3pt}
  \renewcommand{\arraystretch}{1.12}
  \resizebox{\linewidth}{!}{%
  \begin{tabular}{@{}l cccc@{}}
  \toprule
  & & \multicolumn{3}{c}{\textsc{Math}} \\
  \cmidrule(lr){3-5}
  Policy & Avg. & AIME 25 & AIME 26 & HMMT-Feb \\
  \midrule
  \multicolumn{5}{@{}>{\columncolor{tgopdGroupTint}[0pt][0pt]}l@{}}{\emph{Qwen3.5-4B}} \\[2pt]
  Vanilla OPD    & 62.20 & 61.1 & 71.2 & 54.3 \\
  GRPO fallback  & \textbf{64.67} & \textbf{64.8} & \textbf{73.5} & \textbf{55.7} \\
  Mask only      & 63.52 & 62.3 & 72.8 & 55.5 \\
  \midrule
  \multicolumn{5}{@{}>{\columncolor{tgopdGroupTint}[0pt][0pt]}l@{}}{\emph{Qwen3.6-35B-A3B}} \\[2pt]
  Vanilla OPD    & 71.60 & 73.4 & 77.3 & 64.1 \\
  GRPO fallback  & \textbf{72.27} & \textbf{74.0} & \textbf{78.7} & 64.1 \\
  Mask only      & 71.67 & 73.5 & 77.2 & \textbf{64.3} \\
  \bottomrule
  \end{tabular}%
  }
\end{minipage}
\end{lrbox}

\subsection{Ablation: What the Gate Does After It Closes}
\label{ssec:fallback}

\begin{wraptable}{r}{0.56\linewidth}
  \centering
  \vspace{-8pt}
  \usebox{\tgopdFallbackTable}
\end{wraptable}

The threshold ablation examines when teacher supervision should be admitted.
We next ask whether the benefit of gating depends on the update used after
the gate closes. Our main TGOPD configuration uses verifier-grounded GRPO
on audit-rejected prompts. We compare it with a mask-only variant that
suppresses the teacher signal without applying a replacement policy-gradient
update. Both variants use the same gate configuration
($K_T=3$, $\tau=2/3$).
This comparison tests whether withholding teacher supervision is sufficient
to improve performance and whether the GRPO fallback provides an additional
benefit.

Table~\ref{tab:fallback-sopd} reports the single-domain mathematics runs,
using the same Vanilla OPD and TGOPD results as Table~\ref{tab:main}.
Both closed-gate policies improve the average over Vanilla OPD at both scales.
At 4B, GRPO fallback reaches $64.67$, compared with $63.52$ for masking
and $62.20$ for Vanilla OPD; the gains over Vanilla OPD are $2.47$ and
$1.32$ points, respectively.
At 35B, the corresponding averages are $72.27$, $71.67$, and $71.60$,
with gains of $0.67$ and $0.07$ points.
The fallback exceeds masking on all three 4B benchmarks and on both
35B AIME benchmarks. Masking retains a small advantage on 35B HMMT-Feb
($64.3$ vs.\ $64.1$).

To examine whether audit-rejected mathematics prompts retain a learning signal,
we conduct an auxiliary analysis of matched student rollout groups.
Mixed-outcome groups account for $79.0\%$ and $84.7\%$ at 4B and 35B,
respectively, indicating that these prompts can still yield non-zero
group-centered advantages.
Appendix~\ref{app:gated-off-signals} details the diagnostic setup and results.

\WFclear
\suppressfloats[t]
\begin{table}[!htbp]
\centering
\caption{Closed-gate policy on multi-domain runs at 199 training steps.
All three rows within a block share a training corpus and student initialization; the two
gated rows additionally share the gate ($K_T\!=\!3$, $\tau\!=\!2/3$) and differ only in the
update applied to gated-off prompts.
\textbf{Bold}: best per column within a block at the displayed precision,
including ties. Avg.\ is computed before display rounding.}
\label{tab:fallback-mopd}
\small
\setlength{\tabcolsep}{4pt}
\renewcommand{\arraystretch}{1.12}
\resizebox{\linewidth}{!}{%
\begin{tabular}{@{}l c ccc cc cc@{}}
\toprule
& & \multicolumn{3}{c}{\textsc{Math}}
& \multicolumn{2}{c}{\textsc{Code}}
& \multicolumn{2}{c}{\textsc{IF}} \\
\cmidrule(lr){3-5}\cmidrule(lr){6-7}\cmidrule(lr){8-9}
Closed-gate policy & Avg.
  & AIME 25 & AIME 26 & HMMT-Feb
  & LCB & OJBench
  & IFBench & IFEval \\
\midrule
\multicolumn{9}{@{}>{\columncolor{tgopdGroupTint}[0pt][0pt]}l@{}}{\emph{Qwen3.5-4B}} \\[2pt]
Vanilla MOPD \;(no gate)
  & 53.40 & 58.1 & 63.8 & 50.7
  & 50.5 & 18.1 & 48.8 & 83.9 \\
GRPO fallback
  & 54.54 & 58.1 & 67.6 & 51.3
  & 50.0 & \textbf{19.0} & \textbf{51.7} & \textbf{84.1} \\
Mask only
  & \textbf{54.97} & \textbf{60.0} & \textbf{69.8} & \textbf{52.0}
  & \textbf{50.8} & 18.5 & 50.9 & 82.9 \\
\midrule
\multicolumn{9}{@{}>{\columncolor{tgopdGroupTint}[0pt][0pt]}l@{}}{\emph{Qwen3.6-35B-A3B}} \\[2pt]
Vanilla MOPD \;(no gate)
  & 60.99 & 73.1 & 77.3 & 60.7
  & 57.4 & 23.3 & 44.4 & 90.7 \\
GRPO fallback
  & \textbf{61.94} & 73.0 & \textbf{78.0} & 62.4
  & \textbf{57.8} & \textbf{27.6} & 44.0 & \textbf{90.8} \\
Mask only
  & 61.78 & \textbf{74.2} & 76.8 & \textbf{62.5}
  & 57.1 & 27.2 & \textbf{44.6} & 90.1 \\
\bottomrule
\end{tabular}
}
\end{table}

Table~\ref{tab:fallback-mopd} repeats the comparison under multi-domain training, where the
seven-benchmark average summarizes how each policy behaves across domains. The ordering
changes with scale: masking is stronger at 4B ($+1.57$ vs.\ $+1.14$), leading on all three
mathematics columns and LCB, while the fallback recovers the lead at 35B ($+0.95$ vs.\
$+0.79$). The clearest domain-level pattern appears in code: the GRPO fallback outperforms masking
on OJBench at both scales ($19.0$ vs.\ $18.5$; $27.6$ vs.\ $27.2$), where
execution-based verification provides a direct correctness signal.

\paragraph{Gating is effective under both closed-gate policies.}
Both masking and the GRPO fallback improve the average benchmark score
over the corresponding ungated baseline in all four evaluated settings.
The additional benefit of the fallback is setting-dependent: relative to
masking, it changes the average score by $+1.15$ and $+0.60$ points on
single-domain mathematics at 4B and 35B, and by $-0.43$ and $+0.16$ points
under multi-domain training.
These results support prompt-level teacher admission as the central
mechanism, while showing that a verifier-grounded replacement update
is not uniformly better than masking.
We use GRPO as the closed-gate policy in our main experiments, but the
ablation indicates that the choice of closed-gate update remains an
important design consideration.

\WFclear

\vspace{-0.5em}
\section{Related Work}
\label{sec:related}

\subsection{On-Policy Distillation and Multi-Teacher Frameworks}

On-policy distillation (OPD) trains a student on its own rollouts while a stronger teacher
provides token-level supervision through a mode-seeking reverse-KL objective
\citep{gu2024minillm,agarwal2024gkd,opdsurvey2026}. Compared with a single sequence-level
reward, this dense feedback can substantially improve sample efficiency
\citep{thinkingmachines2025opd}. Vanilla OPD applies teacher supervision uniformly across
prompts, while multi-teacher OPD routes each prompt to a domain-expert teacher to integrate
multiple domain capabilities into one student
\citep{mimo2026v2flash,mopd2026,sun2026d}.

Recent analyses show that teacher usefulness varies across instances. OPD may fail when
teacher knowledge is not locally exploitable by the student
\citep{rethinkingopd2026}, when student prefixes drift from teacher-supported states
\citep{revisitingopd2026}, or when an overly strong teacher induces negative transfer into
a smaller student \citep{smallmodels2025}. Existing multi-teacher systems route prompts by
domain but do not verify the selected teacher on each instance before using its token-level
rewards. These studies motivate instance-aware teacher selection, but do not estimate
prompt-level reliability from repeated verifier-scored teacher rollouts. TGOPD performs
this verification before admitting dense teacher supervision.

\subsection{Reliability-Aware Distillation and the Confidence--Correctness Gap}

One line of reliability-aware distillation adapts OPD using signals derived from model
distributions. EOPD introduces forward KL at high-entropy teacher tokens
\citep{eopd2026}, while TrOPD defines token-level trust regions using teacher--student
decoding agreement \citep{tropd2026}. REOPOLD stabilizes implicit token rewards through
mixture clipping and student-entropy sampling \citep{reopold2026}, whereas REOPD
extrapolates token rewards from compatibility signals under a batch-level budget, without
an external verifier \citep{reopd2026}. When no external grader is available, GATES uses
agreement among sampled tutor traces as a trajectory-level reliability proxy
\citep{gates2026}. These methods address uncertainty, diversity, distribution mismatch, or
optimization stability, but do not directly verify outcome correctness. PW-OPSD further
shows that high teacher entropy can reflect either non-viable uncertainty or benign
solution diversity \citep{pwopsd2026}, while ESR finds that the position-dependent
degradation of off-policy teacher guidance is not fully explained by KL or entropy
\citep{esr2026}.

A complementary line explicitly incorporates outcome evidence. RG-OPD conditions
trajectory-level teacher supervision on agreement between verifier feedback and the
teacher--student likelihood gap \citep{rgopd2026}. RLSD uses environmental correctness to
determine the update direction while self-distillation modulates its magnitude
\citep{rlsd2026}. At a finer granularity, SPOT evaluates teacher-proposed candidates using
verifier-scored student continuations and distills outcome-calibrated local targets
\citep{spot2026}. Related data-selection methods verify static responses or tokens before
training \citep{dong2023raft,singh2024restem,xu2025speculativekd}. TGOPD differs in the
granularity and timing of its decision: it verifies repeated complete teacher rollouts to
estimate prompt-level reliability before dense supervision is admitted. Each prompt is
then routed exclusively to either OPD or GRPO, rather than blending the two signals or
calibrating individual trajectories, tokens, or candidate branches.

\section{Conclusion}
\label{sec:conclusion}

This paper addresses vanilla OPD's unconditional trust in teacher supervision through
Teacher-Gated On-Policy Distillation (TGOPD), a prompt-level gate grounded in
verifier-scored teacher probes. TGOPD admits dense OPD only when the reliability audit
passes and otherwise routes the prompt to verifier-grounded GRPO, thereby withholding
teacher supervision selectively rather than weakening it on every prompt. Across 4B and
35B students in mathematics, code, and instruction following, TGOPD outperforms vanilla OPD
in all six single-domain settings and increases the seven-benchmark average by $1.14$ and
$0.95$ points in the corresponding multi-domain settings. The probes largely reuse idle
teacher capacity, raising utilization from $9.8\%$ to $78.9\%$ in the measured 4B
single-domain run; a matched 35B code run incurs a $5.9\%$ increase in mean step time. The
current method requires an automatic verifier and makes a binary routing decision, so extending
outcome-grounded reliability estimation to open-ended tasks and uncertainty-aware gates is
an important direction for future work.

\bibliography{bibliography/references}
\bibliographystyle{styles/colm2026_conference}

\clearpage
\appendix
\raggedbottom
\setlength{\parskip}{4pt}

\section{Contributors}
\label{app:contributors}

\noindent Authors are listed in order of contribution.

\vspace{4pt}
\begingroup
\renewcommand{\thefootnote}{\fnsymbol{footnote}}
\noindent
Zhiwei Zhang\footnotemark[1],
Zechen Sun\footnotemark[1],
Fei Zhao\footnotemark[2],
Kang Peng,
Bin Liang,\\[3pt]
Huayu Deng,
Yao Hu,
Kam-Fai Wong\textsuperscript{\faEnvelope[regular]},
Mu Chuan\textsuperscript{\faEnvelope[regular]}.
\footnotetext[1]{Equal contribution.}
\footnotetext[2]{Project lead.}
\begingroup
\renewcommand{\thefootnote}{\protect\faEnvelope[regular]}
\footnotetext[3]{Corresponding authors.}
\endgroup
\endgroup

\section{Training Hyperparameters and Infrastructure}
\label{app:training-detail}

Table~\ref{tab:training-detail} reports the settings retained in the run records at both
model scales. Within each controlled comparison, Vanilla OPD and TGOPD share the student
initialization, data, optimization settings, and system topology. IcePop/TIS is an
implementation-level safeguard for the asynchronous engine; the method itself is defined
by the gate and routed advantage in \eqref{eq:tgopd-adv}.

\begin{table}[H]
\centering
\caption{Training configuration by model scale. Both scales follow the same optimization
and online-probing protocol; only model-dependent system settings differ.}
\label{tab:training-detail}
\footnotesize
\setlength{\tabcolsep}{4pt}
\renewcommand{\arraystretch}{1.10}
\begin{tabular}{@{}>{\raggedright\arraybackslash}p{\dimexpr0.26\linewidth-8pt\relax}
                    >{\raggedright\arraybackslash}p{\dimexpr0.37\linewidth-4pt\relax}
                    >{\raggedright\arraybackslash}p{\dimexpr0.37\linewidth-4pt\relax}@{}}
\toprule
\multicolumn{1}{@{}>{\columncolor{tgopdGroupTint}[0pt][\tabcolsep]}p{\dimexpr0.26\linewidth-8pt\relax}}{Setting} &
\multicolumn{1}{>{\columncolor{tgopdGroupTint}}p{\dimexpr0.37\linewidth-4pt\relax}}{Qwen3.5-4B} &
\multicolumn{1}{>{\columncolor{tgopdGroupTint}[\tabcolsep][0pt]}p{\dimexpr0.37\linewidth-4pt\relax}@{}}{Qwen3.6-35B-A3B} \\
\midrule
Runtime & slime asynchronous stack & slime asynchronous stack \\
Execution & Asynchronous rollout and update & Asynchronous rollout and update \\
Rollout batch size & 64 prompts per cycle & 64 prompts per cycle \\
Responses per prompt ($G$) & 4 & 4 \\
Update batch size & 128 & 128 \\
Updates per rollout cycle & 2 & 2 \\
Optimizer & Adam ($\beta_1=0.9$, $\beta_2=0.98$, $\epsilon=10^{-8}$) &
  Adam ($\beta_1=0.9$, $\beta_2=0.98$, $\epsilon=10^{-8}$) \\
Learning rate & $1\times10^{-6}$ & $1\times10^{-6}$ \\
Learning-rate schedule & Constant & Constant \\
Weight decay & 0.1 & 0.1 \\
Advantage estimator & Group mean, without std normalization &
  Group mean, without std normalization \\
Policy surrogate & PPO clipping, $\epsilon=0.2$ & PPO clipping, $\epsilon=0.2$ \\
Asynchronous mismatch control & IcePop/TIS, ratio clip 2.0 &
  IcePop/TIS, ratio clip 2.0 \\
Probe realization & Three fresh online teacher rollouts &
  Three fresh online teacher rollouts \\
Teacher probes ($K_T$) & 3 & 3 \\
Default reliability threshold ($\tau$) & $2/3$ & $2/3$ \\
Maximum response length & 16,384 (math); 8,192 (code and IF) &
  16,384 (math); 8,192 (code and IF) \\
\bottomrule
\end{tabular}
\end{table}

\paragraph{Infrastructure.}
The 4B runs use five 8-GPU nodes (40 GPUs): two trainer nodes, two
student-rollout nodes, and one teacher node. The 35B runs use seven
8-GPU nodes (56 GPUs): four trainer nodes, two student-rollout nodes, and one
teacher node. For the 4B 16K runs, we cap the SGLang static memory fraction at 0.60 and
both the maximum running requests and teacher-server concurrency at 64.

\paragraph{Probe realization.}
At both model scales, the frozen teacher draws three fresh probe responses online for each
prompt and the gate opens when at least two pass the verifier. Thus the 4B and 35B
experiments instantiate the same reliability test described in \S\ref{sec:method:gate};
the resource topology changes with model scale, but the training rule does not.

\clearpage
\section{Evaluation Benchmark Details}
\label{app:eval-detail}

Table~\ref{tab:eval-detail} lists the benchmarks used for in-domain evaluation, together
with the number of independent generation runs and the metric reported in the main tables.
For benchmarks evaluated with multiple runs, the reported score is the average over those
runs. For LiveCodeBench we report only the code-generation subtask; for OJBench we report
the overall score aggregated over the C++ and Python splits.

\begin{table}[H]
\centering
\caption{Evaluation benchmark details.
\emph{Runs}: number of independent generations per checkpoint; the score reported in all
main-text tables is the mean over these runs.}
\label{tab:eval-detail}
\small
\renewcommand{\arraystretch}{1.05}
\begin{tabular}{@{}
  p{\dimexpr0.13\linewidth-1.5\tabcolsep\relax}
  p{\dimexpr0.42\linewidth-1.5\tabcolsep\relax}
  >{\centering\arraybackslash}p{\dimexpr0.10\linewidth-1.5\tabcolsep\relax}
  p{\dimexpr0.35\linewidth-1.5\tabcolsep\relax}@{}}
\toprule
Domain & Benchmark & \#Runs & Metric \\
\midrule
\multirow{3}{*}{Math}
  & AIME 2025  & 64 & Accuracy (avg64) \\
  & AIME 2026  & 64 & Accuracy (avg64) \\
  & HMMT-Feb 2025 & 32 & Accuracy (avg32) \\
\addlinespace[3pt]
\multirow{2}{*}{Code}
  & LiveCodeBench (code generation) & 6 & Pass@1 (avg6) \\
  & OJBench (overall)               & 1 & Aggregate (C++/Python) \\
\addlinespace[3pt]
\multirow{2}{*}{IF}
  & IFBench & 1 & Accuracy \\
  & IFEval  & 1 & Accuracy \\
\bottomrule
\end{tabular}
\end{table}

\section{Per-Configuration GPU Utilization Traces}
\label{app:gpu-traces}

\begin{figure}[H]
\centering
\includegraphics[width=0.64\linewidth]{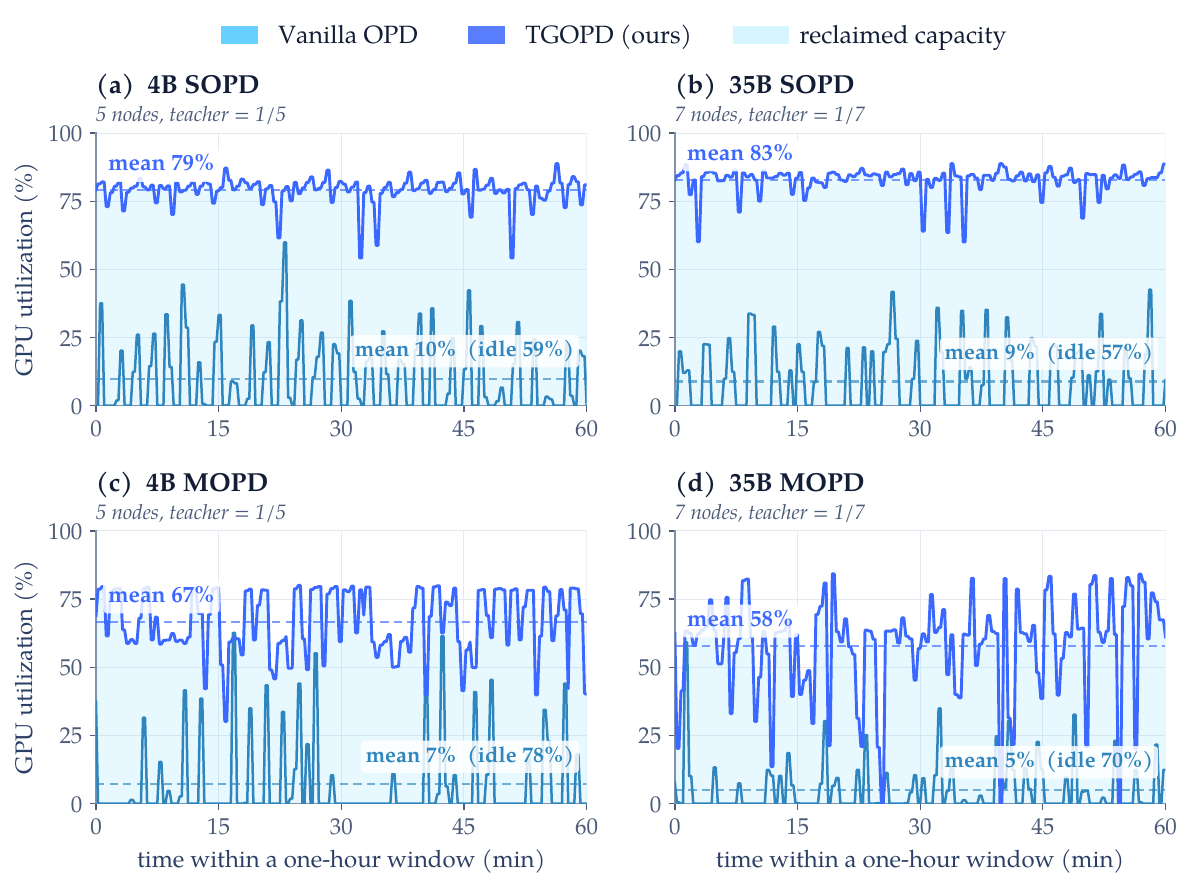}
\caption{\textbf{Teacher-node GPU utilization over a one-hour window for each configuration.}
Light blue: vanilla OPD. Dark blue: TGOPD. The shaded region indicates the teacher capacity
reclaimed through probe generation. Dashed lines indicate the mean utilization of each run.
\textbf{(a,b)} Single-domain OPD at the 4B and 35B scales.
\textbf{(c,d)} Multi-domain OPD at the corresponding scales.}
\label{fig:gpu-traces}
\end{figure}

Section~\ref{ssec:gpu-util} summarizes teacher-node utilization by its mean and idle
fraction (Table~\ref{tab:gpu-util}) and its marginal distribution
(Figure~\ref{fig:gpu-util}a). Neither summary preserves the time axis.
Figure~\ref{fig:gpu-traces} therefore plots the raw one-hour trace for every configuration.
All eight traces are sampled at 15-second resolution over a one-hour window taken from
steady-state training, and the two curves within a panel are drawn on a common axis.

\paragraph{Matched end-to-end timing.}
To avoid inferring wall-clock neutrality from utilization alone, we additionally compare
the completed 35B CodeIO TGOPD and OPD runs under identical settings. Across $500$ aligned
rollout cycles, TGOPD increases mean step time by $5.9\%$, while mean decode throughput
changes by less than $0.1\%$. Thus most probe work overlaps with idle teacher capacity,
but the audit is not strictly free.

The traces clarify three aspects of the utilization results in \S\ref{ssec:gpu-util}.

\paragraph{The baseline teacher is bursty, not uniformly slow.}
In all four panels the light-blue curve alternates between brief spikes and long
stretches pinned at zero, rather than hovering at a low but steady level. Each spike is one scoring
pass; the flat intervals are the student's rollout phase, during which the teacher has no
scoring work. The idle fraction in Table~\ref{tab:gpu-util} measures these intervals. Their
duration is set by the student rollout rather than teacher scoring speed, so shrinking the
teacher or batching its scoring work differently would not remove the gap.

\paragraph{Probes occupy otherwise low-utilization intervals.}
The TGOPD traces show fewer near-zero utilization intervals than
the OPD traces, consistent with the teacher performing additional
probe generation between scoring workloads.
The traces describe when the teacher is active, but do not isolate
queueing delays or competition between probing and scoring.
We therefore assess runtime impact using the matched timing comparison
rather than inferring it from utilization peaks.

\paragraph{Multi-domain probing oscillates because it is domain-routed.}
The SOPD panels show a tight blue band around $79$--$83\%$, whereas the MOPD panels swing
between roughly $40\%$ and $85\%$. The MOPD teacher node is partitioned into three
domain-specialist engines (code 4\,GPU, math 2\,GPU, IF 2\,GPU). Each prompt activates
only its corresponding teacher, while different prompts can use different teachers
concurrently. Node occupancy therefore varies with the domain mix of the batch. This oscillation is what produces the bimodal marginal
distribution visible in Figure~\ref{fig:gpu-util}a and the lower MOPD means in
Table~\ref{tab:gpu-util}. It reflects how the node is divided rather than a failure of the
probe to fill the window: the idle fraction still falls to $0$--$2\%$.

\section{Reward Variation on Audit-Rejected Mathematics Prompts}
\label{app:gated-off-signals}

To complement Table~\ref{tab:fallback-sopd}, we examine student reward
variation on mathematics prompts for which the teacher gate closes.
The analysis follows the main training configuration, with three online
teacher probes per prompt and four responses per student group.

We classify each recorded student group as mixed-outcome, all-correct, or
all-incorrect, and aggregate group counts over the rejected prompts. Repeated
groups from different training steps are included. Under binary rewards,
mixed-outcome groups contain both correct and incorrect responses and therefore
yield non-zero group-centered advantages, whereas all-correct and all-incorrect
groups yield zero advantages.

\begingroup
\setlength{\intextsep}{6pt}
\begin{figure}[!htbp]
\centering
\includegraphics[width=0.78\linewidth]{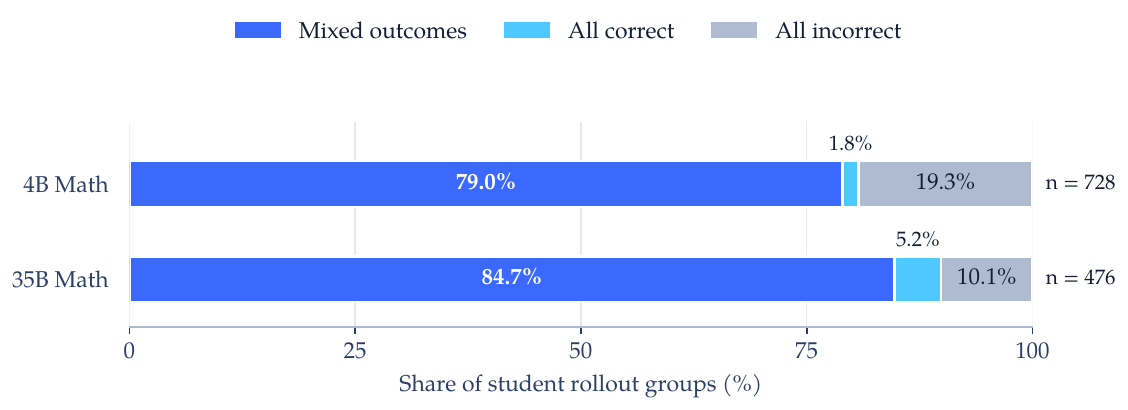}
\small
\caption{\textbf{Student reward variation on audit-rejected mathematics prompts.}
Mixed-outcome groups yield non-zero GRPO advantages, whereas all-correct and
all-incorrect groups yield zero advantages under binary rewards. Percentages are
computed over recorded rollout groups, and $n$ denotes the number of rejected
prompts. The analysis follows the main training configuration with four student
responses per group.}
\label{fig:gated-off-math-signal}
\end{figure}
\endgroup

Figure~\ref{fig:gated-off-math-signal} shows that mixed-outcome groups account
for $79.0\%$ and $84.7\%$ at 4B and 35B, respectively. All-incorrect groups
account for $19.3\%$ and $10.1\%$, while all-correct groups account for
$1.8\%$ and $5.2\%$. Thus, withholding teacher supervision does not
necessarily eliminate within-group reward variation among student rollouts.
This finding complements the ablation results reported in
Section~\ref{ssec:fallback} by showing that audit-rejected prompts can
still provide non-zero verifier-grounded advantages.

\end{document}